\pdfoutput=1

\documentclass[11pt]{article}

\usepackage{acl}
\usepackage{placeins}
\usepackage{times}
\usepackage{latexsym}
\usepackage{float}
\usepackage{amsmath,amssymb}
\usepackage[T1]{fontenc}
\usepackage{xcolor}
\usepackage{tabularx}
\usepackage{booktabs}

\usepackage[utf8]{inputenc}

\usepackage{microtype}

\usepackage{inconsolata}

\usepackage{graphicx}

\usepackage{multirow}
\usepackage{arydshln} 
\usepackage{booktabs} 
\usepackage{caption} 
\usepackage{natbib} 
\usepackage{hyperref} 
\usepackage{enumitem}
\usepackage{linguex}
\makeatletter

\title{Evaluating Pragmatic Reasoning in Large Language Models:\\
Evidence from Scalar Diversity}

\author{Ye-eun Cho\\
  Sungkyunkwan University\\
  Seoul, South Korea\\
  \texttt{joyenn@skku.edu}}
\begin{document}
\maketitle
\begin{abstract}
Evaluating pragmatic reasoning in large language models (LLMs) remains challenging because model behavior can vary depending on evaluation methods. Previous studies suggest that prompt-based judgments may diverge from models’ internal probability distributions, raising questions about whether observed performance reflects underlying competence or task-induced behavior. This study examines this issue using scalar diversity as a graded diagnostic for pragmatic inference. Following \citet{hu2023prompting}, this study compares direct probability measurement and metalinguistic prompting across multiple models and experimental settings. The results show that neither evaluation method consistently outperforms the other and that pragmatic behavior varies substantially across model families, prompting strategies, and task structures. Moreover, scalar diversity gradients emerge only in specific model–condition combinations, suggesting that pragmatic reasoning in LLMs reflects an interaction between internal probabilistic representations and task-induced prompting behavior rather than a stable competence captured by a single evaluation paradigm. These findings highlight the central role of evaluation design in interpreting pragmatic abilities in LLMs.\\

\end{abstract}
\section{Introduction}

\begin{figure}[t]
  \includegraphics[width=\columnwidth]{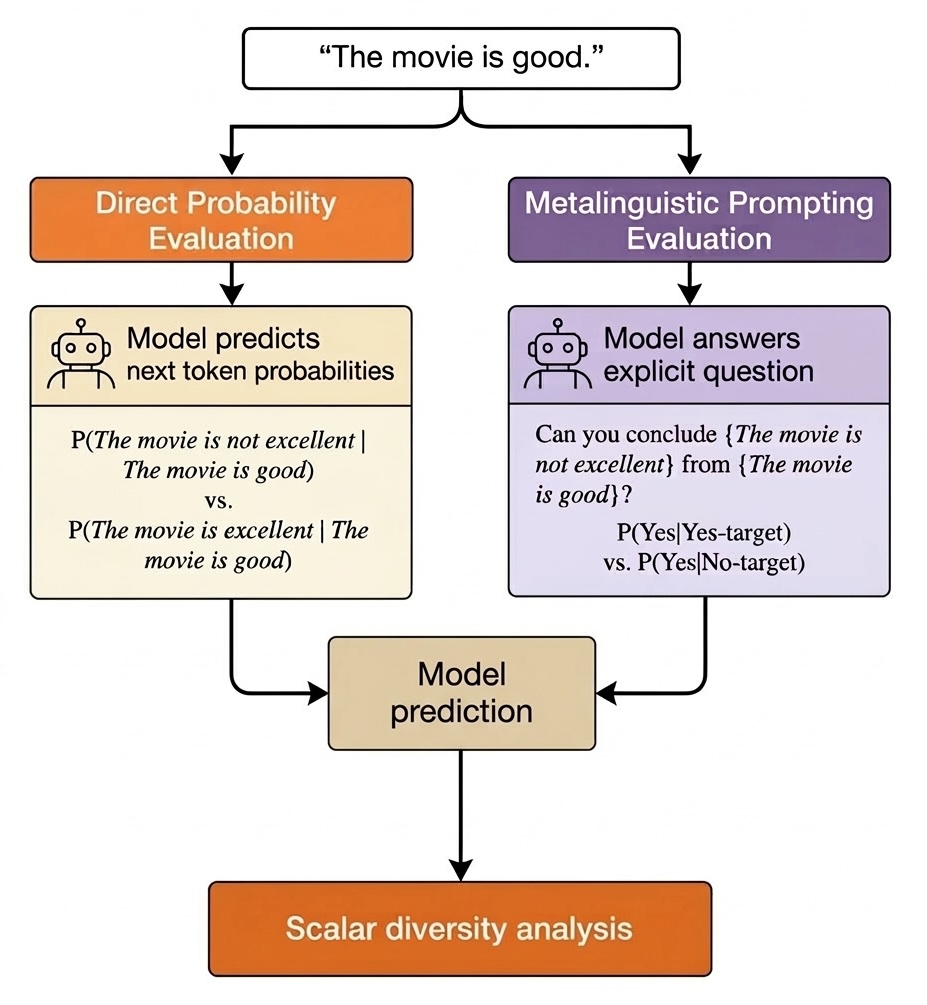}
  \caption{Overview of the two evaluation paradigms used in this study}
  \label{fig:fig1}
\end{figure}

Recent advances in large language models (LLMs) have demonstrated remarkable performance across a wide range of linguistic tasks \citep{wang2019superglue, hu2024auxiliary, marjieh2024large}. Despite these successes, however, LLMs continue to show notable limitations in the domain of pragmatics, particularly in their ability to reason about implicit meaning \citep{mielke2022reducing, webson2022prompt, turpin2023language, cong2024manner}. Interestingly, while many studies report limited pragmatic abilities in LLMs \citep{webson2022prompt, mielke2022reducing, turpin2023language, cong2024manner}, other experiments have shown that certain prompt designs or task framings can substantially improve model performance on pragmatic reasoning tasks \citep{cong2024manner, cho-maeng-2025-vision, cho2025prompting, shulginov2025evaluating}. This raises a fundamental question: do LLMs genuinely possess pragmatic competence, or do they merely appear competent under specific evaluation conditions?

To address this issue, the present study adopts the competence–performance framework proposed by \citet{chomsky1965aspects}. Competence refers to internalized linguistic knowledge, whereas performance reflects language use under particular cognitive and contextual conditions. \citet{hu2023prompting} applied this distinction to LLM evaluation, emphasizing the importance of separating underlying model knowledge from prompt-induced behavior. However, their work primarily focused on local linguistic phenomena, leaving pragmatic abilities largely unexplored.

We investigate this question through the phenomenon of scalar implicature, a widely studied form of pragmatic inference. Scalar implicature arises when, on the lexical scale <\textit{some, all}>, a weaker term (e.g., \textit{some}) leads to the inference that a stronger alternative (e.g., \textit{all}) does not hold \citep{horn1972semantic, grice1975logic, levinson2000presumptive}. Importantly, the strength of such implicatures varies across lexical scales, a phenomenon known as scalar diversity \citep{van2016scalar}. For example, <\textit{some, all}> reliably triggers implicatures in human interpretation, whereas pairs such as <\textit{warm, hot}> often yield weaker inferences. This variability provides a useful diagnostic for evaluating models’ sensitivity to pragmatic reasoning.

Following the evaluation framework proposed by \citet{hu2023prompting}, the present study examines LLMs’ interpretation of scalar implicatures using two evaluation methods: direct probability measurement and metalinguistic prompting, as illustrated in Figure~\ref{fig:fig1}. In addition, we investigate how prompting strategies and model size influence the observed patterns.

\section{Background}
\subsection{Direct vs. Metalinguistic Evaluation}
\citet{hu2023prompting} systematically compared two evaluation methods for assessing the linguistic knowledge of LLMs: direct probability measurement and metalinguistic prompting. Direct probability measurements evaluate LLMs by directly accessing their internal probabilistic representations, such as next-token probabilities or sentence-level pseudo-likelihoods. These probabilities are often interpreted as the most transparent reflection of what a model “knows,” derived from its learned representations. In contrast, metalinguistic prompting presents models with natural-language queries that require them to make explicit judgments about linguistic stimuli. This approach asks the model to externalize its knowledge through prompted responses.

Using these two evaluation methods, \citet{hu2023prompting} conducted a series of experiments across multiple linguistic domains, including word prediction, word comparison, and grammaticality judgment. They implemented one direct probability-based method and three metalinguistic prompting methods that varied in their structural similarity to the direct format. For instance, MetaQuestionSimple closely mirrors the direct method by placing the prediction target immediately adjacent to the query, whereas MetaInstruct introduces an instructional prompt format and MetaQuestionComplex embeds the target in a more elaborate context.

Their experiments revealed several key findings. First, metalinguistic judgments often diverge from direct probability measurements, indicating that models’ explicit responses may not faithfully reflect their internal representations. Second, direct probability measurements generally outperform metalinguistic prompting in linguistic evaluation tasks. Third, presenting sentences as minimal pairs improves metalinguistic performance compared to evaluating sentences in isolation. Finally, the consistency between direct and metalinguistic responses decreases as the prompting format becomes more distant from the direct probability structure.

Based on these findings, \citet{hu2023prompting} argued that LLM evaluation should distinguish between competence, reflected in internal probability distributions, and performance, reflected in prompt-based responses. From this perspective, models may possess linguistic knowledge that is not reliably expressed through metalinguistic prompts.

Subsequent study has extended this discussion to pragmatic reasoning. For example, \citet{cong2024manner} evaluates manner implicature using both probability-based and prompting-based methods, but does not directly compare these approaches within a unified model framework, limiting the comparability of the results. To address this limitation, the present study applies the competence–performance framework to scalar implicatures by systematically comparing direct probability measurements and metalinguistic prompting within the same models.

\begin{table*}[t]
\centering
\small
\setlength{\tabcolsep}{4pt}
\resizebox{\textwidth}{!}{
\begin{tabular}{llp{3.2cm}p{3.5cm}p{3.5cm}}
\toprule
\textbf{Weak} & \textbf{Strong} & \textbf{Anchor Sentence} & \textbf{Yes-target} & \textbf{No-target} \\
\midrule
Good & Excellent & \textit{The movie is good} & \textit{The movie is not excellent} & \textit{The movie is excellent} \\
Warm & Hot & \textit{The soup is warm} & \textit{The soup is not hot} & \textit{The soup is hot}\\
\ldots & \ldots & \ldots & \ldots & \ldots \\
\bottomrule
\end{tabular}
}
\caption{Sample items from the scalar diversity evaluation dataset}
\label{tab:tab1}
\end{table*}

\subsection{Scalar Implicature}
Scalar implicature is a well-studied form of pragmatic inference in which the use of a weaker expression on a lexical scale leads the listener to infer that a stronger alternative does not hold \citep{horn1972semantic, grice1975logic, levinson2000presumptive}. Consider the example sentence (1).

\ex.Some cookies were eaten.

From a logical perspective, the quantifier \textit{some} means \textit{at least one and possibly all}. Therefore, the literal meaning of (1) is compatible with both the interpretation that \textit{only one cookie was eaten} and the interpretation that \textit{all cookies were eaten}. Nevertheless, in ordinary communication, listeners often infer (1) as \textit{not all cookies were eaten}. This inference arises because the speaker chose the weaker expression (\textit{some}) despite the availability of the more informative alternative (\textit{all}), leading the listener to infer that the stronger alternative does not hold; otherwise, the speaker would have used the stronger expression instead.

Recent experimental research has shown that scalar implicatures do not occur uniformly across lexical scales. Instead, different scalar pairs vary in how strongly implicatures are inferred, a phenomenon known as scalar diversity \citep{van2016scalar}. Consider the examples in (2) and (3).

\ex.
\a. John ate some cookies.
\b. John did not eat all cookies.

\ex.
\a. The coffee is warm.
\b. The coffee is not hot.

Listeners readily infer the meaning in (2b) from the sentence in (2a), interpreting \textit{some} as implying \textit{not all}. In contrast, the inference from \textit{warm} in (3a) to \textit{not hot} in (3b) is much weaker and less consistently derived, illustrating that scalar implicatures vary in strength across different lexical scales (\citealp{van2016scalar, ronai2021pragmatic, ronai2024could, pankratz2021role}; see also \citealp{degen2015processing}).

This variability across scalar pairs provides a useful diagnostic for evaluating pragmatic reasoning. In the context of LLM evaluation, examining multiple scalar pairs rather than a single item allows for a more fine-grained assessment of models’ pragmatic competence.

Previous work on LLMs has often evaluated scalar implicature using a single scalar pair such as <\textit{some, all}> \citep{cho2024pragmatic}. However, focusing on a single item may overestimate models’ pragmatic competence. To address this limitation, the present study adopts a scalar diversity framework for assessing models' pragmatic reasoning abilities.

\definecolor{softgray}{gray}{0.4}
\definecolor{softred}{RGB}{180,60,60}
\definecolor{softblue}{RGB}{60,100,180}
\begin{table*}[t]
\centering
\small
\setlength{\tabcolsep}{5pt}

\begin{tabularx}{\textwidth}{lX}
\toprule
\bfseries Evaluation & \bfseries Example \\
\midrule

Direct &
\{\textcolor{softgray}{The movie is good}, \textcolor{softred}{The movie is not excellent}\} \\

MetaSimple &
Can you conclude from \{\textcolor{softgray}{The movie is good}\} that \{\textcolor{softred}{The movie is not excellent}\}? Respond with either Yes or No as your answer. \\

MetaInstruct &
You are a helpful writing assistant. Tell me if you can conclude from \{\textcolor{softgray}{The movie is good}\} that \{\textcolor{softred}{The movie is not excellent}\}. Respond with either Yes or No as your answer. \\

MetaComplex &
Here is a sentence: \{\textcolor{softred}{The movie is not excellent}\}. Can you conclude this from \{\textcolor{softgray}{The movie is good}\}? Respond with either Yes or No as your answer. Answer: \\

\bottomrule
\end{tabularx}

\caption{Examples of direct probability and metalinguistic prompting in Experiment A}
\label{tab:prompt_examples}
\end{table*}

\begin{table*}[t]
\centering
\small
\setlength{\tabcolsep}{5pt}

\begin{tabularx}{\textwidth}{lX}
\toprule
\bfseries Evaluation & \bfseries Example \\
\midrule

Direct &
\{\textcolor{softgray}{The movie is good}, \textcolor{softred}{The movie is not excellent}\} \\

MetaSimple &
Which sentence can you conclude from \{\textcolor{softgray}{The movie is good}\}?:
1) \{\textcolor{softred}{The movie is not excellent}\}
2) \{\textcolor{softblue}{The movie is excellent}\}.
Respond with either 1 or 2 as your answer. \\

MetaInstruct &
You are a helpful writing assistant. Tell me which sentence you can conclude from
\{\textcolor{softgray}{The movie is good}\}:
1) \{\textcolor{softred}{The movie is not excellent}\}
2) \{\textcolor{softblue}{The movie is excellent}\}.
Respond with either 1 or 2 as your answer. \\

MetaComplex &
Here are two sentences:
1) \{\textcolor{softred}{The movie is not excellent}\}
2) \{\textcolor{softblue}{The movie is excellent}\}.
Which sentence can you conclude from \{\textcolor{softgray}{The movie is good}\}?
Respond with 1 or 2. Answer: \\

\bottomrule
\end{tabularx}

\caption{Examples of direct probability and metalinguistic prompting in Experiment B}
\label{tab:prompt_examples_B}
\end{table*}

\section{Methods}
\subsection{Models}
Following \citet{hu2023prompting}, the Flan-T5 family \citep{chung2024scaling} was included, which is an instruction-tuned encoder–decoder language model. To examine whether the observed patterns generalize beyond this architecture, the Qwen2 family \citep{yang2024qwen2} was additionally evaluated. Qwen models are decoder-only instruction-tuned language models designed for autoregressive next-token prediction. This contrast enables a comparison based on different model architectures.

In addition, multiple model sizes were evaluated within each model family in order to examine potential scaling effects. For Flan-T5, three model sizes were used as: Small, Base, and Large. For Qwen, three parameter scales were evaluated as: 0.5B, 1.5B, and 7B.

\subsection{Materials}
The materials were adapted from the scalar diversity dataset developed by \citet{ronai2024could}, which was originally designed to examine variability in the derivation of scalar implicatures (SIs) across a wide range of lexical scales. In the original dataset, each item consisted of a <weak, strong> scalar pair (e.g., <\textit{good, excellent}>) embedded in a neutral carrier sentence (e.g., \textit{The movie is good}), which could be interpreted either literally or with an SI-enriched meaning.

For the present study, the dataset was expanded using GPT-4o \citep{openai2024gpt4o}. While preserving the original scalar pairs, new carrier sentences were generated in the form “[something] \textit{is good}”, varying the surrounding content while keeping the SI trigger constant. Implausible or identical sentences were filtered and regenerated during the generation process. This procedure produced 100 distinct carrier sentences for each of the 60 scalar pairs, yielding 6,000 sets of items in total.

As exemplified by Table~\ref{tab:tab1}, each evaluation item consists of an anchor sentence together with two candidate interpretations. The Yes-target represents the pragmatic interpretation associated with scalar implicature, while the No-target corresponds to the logical interpretation that does not require pragmatic enrichment. The dataset and experimental code used in this study are publicly available.\footnote{\url{https://github.com/joyennn/pragmatic_reasoning}}

\begin{figure*}[t]
  \centering
  \includegraphics[width=\textwidth]{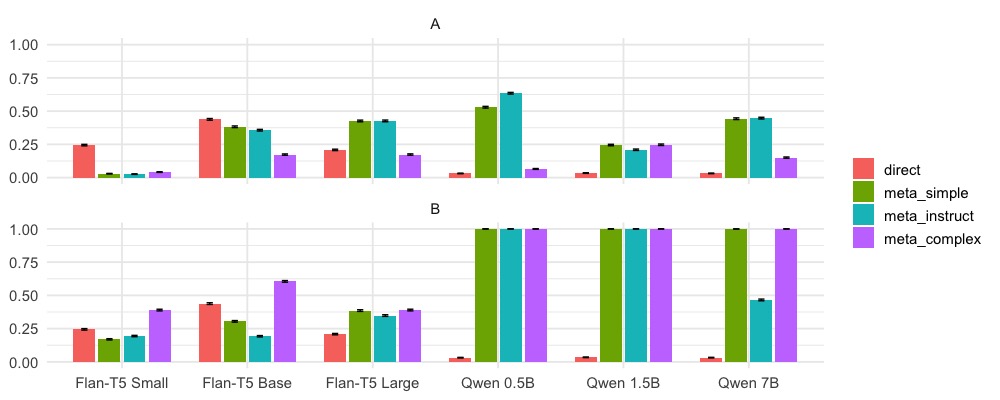}
  \caption{Overall accuracy of scalar inference predictions across models and prompting conditions for Experiments A (top) and B (bottom)}
  \label{fig:overall_accuracy}
\end{figure*}

\subsection{Procedure}
In the experiment, two experimental conditions were used. In Experiment A (sentence judgment), the model evaluates one candidate interpretation at a time and produces an absolute judgment. In Experiment B (sentence comparison), the model is presented with two candidate interpretations simultaneously and must select the preferred interpretation among the two.

In both experimental conditions, model preferences for pragmatic interpretations were evaluated using two paradigms: direct probability and metalinguistic prompting evaluation. For direct probability evaluation, the preference between the two candidate interpretations was estimated by comparing the conditional probabilities of the candidates given the anchor sentence. Specifically, the probability of each candidate continuation was computed as $P(\text{candidate}\mid\text{anchor})$, under the assumption that a continuation that forms a more acceptable interpretation of the anchor sentence will receive a higher conditional probability. For each item, the anchor sentence and each candidate sentence were concatenated and scored by the models. The candidate with the higher conditional probability was taken as the model’s preferred interpretation and coded as TRUE; otherwise, it was coded as FALSE.

For metalinguistic evaluation, three prompting conditions were used: MetaSimple, MetaInstruct, and MetaComplex. Following \citet{hu2023prompting}, these prompt types vary in the distance between the prediction target and the question as well as in the degree of contextual framing. MetaSimple places the target sentence directly within the question. MetaInstruct adds an instructional frame (e.g., “You are a helpful writing assistant”), while MetaComplex embeds the target in a longer contextual frame. The interrogative form “Can you conclude \{\} from \{\}?” was adapted from \citet{ronai2024could} to elicit explicit inferential judgments.

In Experiment A, as in Table~\ref{tab:prompt_examples}, the models were presented with an anchor sentence and a single candidate sentence (either Yes-target or No-target). The model responds with Yes or No. The log-probability of generating the token 'Yes' is computed for both candidates, and the difference between these probabilities is calculated as:
\[
P(\text{Yes}\mid\text{Yes-target}) - P(\text{Yes}\mid\text{No-target})
\]
In Experiment B, as in Table~\ref{tab:prompt_examples_B}, the models receive the anchor sentence together with both candidate sentences simultaneously and must select between 1 (Yes-target) or 2 (No-target). The difference between the log-probabilities of the two responses is computed as:
\[
P(\text{1}) - P(\text{2})
\]
In both experiments, a positive value indicates a preference for the Yes-target and is coded as TRUE. These TRUE responses were used for the subsequent analysis.

Finally, to examine the relationship between direct and metalinguistic evaluations, Pearson correlations were computed between the direct preference scores and the metalinguistic preference scores across items.

\begin{figure*}[t]
  \includegraphics[width=\textwidth]{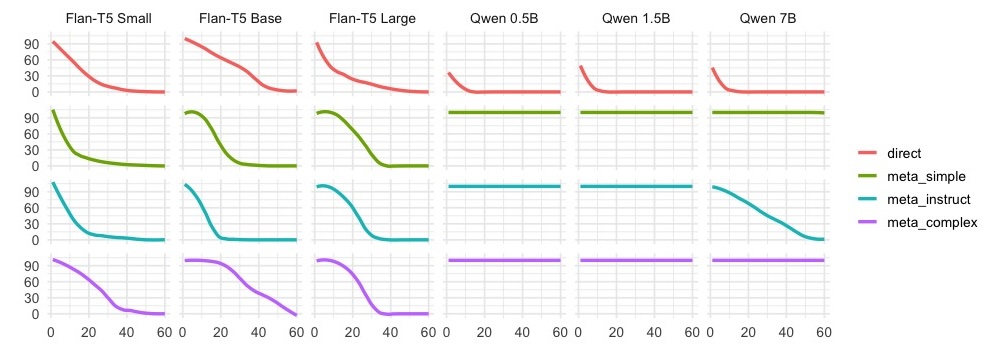}
  \caption{Item-level accuracy across scalar items for each model and evaluation condition in Experiment B}
  \label{fig:fig4}
\end{figure*}

\section{Results}
\subsection{Overall accuracy}
Figure~\ref{fig:overall_accuracy} presents the overall accuracy of scalar inference predictions across models, prompting conditions, and experimental paradigms. Accuracy was calculated as the proportion of TRUE responses across the entire item set. The results reveal systematic differences across prompting strategies, model families, model sizes, and experimental conditions as follows.

\subsubsection{Direct probability vs. Metalinguistic prompting}
First, a clear difference emerges between the direct probability and the metalinguistic prompting conditions. Across nearly all models and experimental settings, the direct probability condition consistently yielded lower accuracy than the metalinguistic prompting conditions. A small number of exceptions are observed in Experiment A for the smaller Flan-T5 models, where direct probability estimates slightly outperform some metalinguistic prompts, but the overall scores remain low.

To examine the relationship between the two evaluation paradigms, Pearson correlations were computed over overall accuracy scores obtained from direct probability estimation and metalinguistic prompting (see Table 4\&5 in Appendix~\ref{app:pearson}). Overall, the correlations were generally weak (typically $r \approx .10-.25$), indicating that the two evaluation strategies produce only partially overlapping patterns of overall accuracy. In several cases, correlations were near zero or even negative, particularly for Flan-T5 Large and some Qwen conditions. These results suggest that direct probability estimation and metalinguistic prompting capture substantially different aspects of model behavior in scalar inference tasks.

\subsubsection{Metalinguistic prompting strategies}
Differences are also observed among the three metalinguistic prompting strategies. While the metalinguistic prompts generally outperform the direct probability condition, the relative performance among the three varies across models and experimental settings. In many cases, the complex prompt (MetaComplex) shows noticeably different accuracy compared to the other two conditions, sometimes performing substantially lower and in other cases slightly higher. However, there are also instances in which all three metalinguistic prompts yield very similar accuracy, as well as occasional idiosyncratic patterns for particular models. Overall, these observations make it difficult to identify a consistent ranking among the three prompting strategies, suggesting that the effectiveness of specific prompt formulations interacts with model characteristics and task conditions.

\begin{figure*}[t]
  \includegraphics[width=\textwidth]{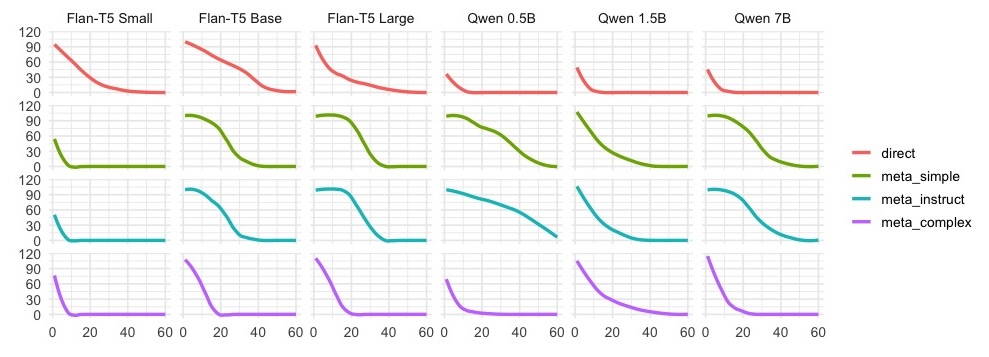}
  \caption{Item-level accuracy across scalar items for each model and evaluation condition in Experiment A}
  \label{fig:fig3}
\end{figure*}

\subsubsection{Model differences}
In addition, clear differences emerge between the two model families, Flan-T5 and Qwen2. While both models show improvements under metalinguistic prompting relative to the direct probability condition, their overall behavioral patterns differ substantially. The Qwen models exhibit a highly polarized pattern: direct probability estimates are consistently very low, whereas metalinguistic prompting often yields extremely high accuracy, in several cases approaching ceiling levels in Experiment B. In contrast, the Flan-T5 models display more moderate patterns in which direct probability estimates remain relatively higher and occasionally even exceed the performance of some metalinguistic prompts.

One possible explanation for these contrasting behaviors lies in differences in model architecture and training objectives. Flan-T5 models are instruction-tuned encoder–decoder models, which may produce probability estimates that remain more sensitive to surface lexical cues in the input sentence. By contrast, Qwen models are decoder-only autoregressive models trained on large-scale instruction-following data, which may respond more strongly to explicit metalinguistic task framing.

\subsubsection{Model size}
Model size effects also differ across the two model families. Within the Flan-T5 family, a generally positive scaling trend can be observed under metalinguistic prompting conditions, with larger models typically achieving higher accuracy than smaller ones. In contrast, the Qwen models do not exhibit a consistent scaling trend. Across several conditions, larger Qwen models do not outperform smaller ones and in some cases even show slightly lower accuracy. This pattern suggests that increasing model size within this family does not systematically improve scalar inference performance.

\subsubsection{Experimental settings}
Moreover, there are clear differences between Experiment A and B. Across nearly all models and prompting conditions, accuracy is substantially higher in Experiment B than in Experiment A. This pattern suggests that the comparison-based evaluation used in Experiment B provides stronger cues for selecting the pragmatically enriched interpretation than the single-sentence judgment task used in Experiment A.

\subsection{Item-level accuracy}
While the overall accuracy analysis provides a useful summary of model performance, it does not directly capture the central phenomenon of scalar diversity. Scalar diversity refers to the graded nature of scalar implicature, where different items exhibit different likelihoods of pragmatic enrichment. That is, some scalar expressions strongly favor the pragmatic interpretation, whereas others do not.

To examine whether models capture this graded tendency, it is necessary to analyze item-level accuracy across scalar items. If models are sensitive to scalar diversity, their predictions should exhibit a gradual pattern across the item spectrum, rather than a flat pattern or an abrupt threshold-like drop. Figures 3 and 4 therefore visualize the gradient patterns across items for each model and prompting condition in Experiments A and B. In addition, the steepness of each gradient was quantified using slope estimates (see Table 6\&7 in Appendix~\ref{app:slopes}), which serve as a useful indicator for capturing the overall tendency of how model predictions vary across scalar items, although they do not provide a definitive measure of gradient structure.

\subsubsection{Overall gradient}
In Figures~\ref{fig:fig3} and~\ref{fig:fig4}, scalar items are ordered according to their tendency to trigger pragmatic interpretations, and model accuracy is plotted across the item spectrum. If models capture this tendency, their predictions should show gradual variation across items.

Across the models, three broad patterns can be observed. First, several conditions show gradual gradients, with slope values typically between approximately $-1$ and $-2$, indicating continuous variation in model predictions across scalar items. Second, some conditions exhibit weaker gradients, with slopes close to zero or within the range of approximately 0 to $-1$, suggesting limited differentiation among scalar expressions. Finally, a small number of conditions produce nearly flat curves, with slopes effectively at zero, indicating that models assign the same interpretation to most items.

\subsubsection{Direct probability vs. Metalinguistic prompting}
In the direct probability condition, several models exhibit relatively weak gradients across scalar items. In particular, the Qwen models show curves that remain relatively flat across the item spectrum ($-0.28$ to $-0.31$) in both experiments. The Flan-T5 models display a somewhat different pattern. Although the curves also show declines across the item spectrum, the slopes are substantially larger in magnitude ($-1.17$ to $-1.90$), indicating stronger changes in predictions across scalar items. In other words, the direct probability condition produces clearer gradients in the Flan-T5 models than in the Qwen models.

Metalinguistic prompting often produces stronger gradients. In many cases, particularly within the Flan-T5 models, the metalinguistic prompts yield slope values between approximately $-1$ and $-2$ or lower, indicating a more pronounced change in predictions across items. However, the effects of metalinguistic prompting are not uniform across models. While some conditions show clearer gradients, others produce nearly constant predictions in the Qwen models, particularly in Experiment B. This variability indicates that the influence of prompting strategies interacts strongly with model architectures and experimental settings.

\subsubsection{Model differences}
Also, Flan-T5 and Qwen model families show substantial differences. Within the Flan-T5 models, the larger models (Base and Large) frequently exhibit stronger gradients across scalar items under metalinguistic prompting conditions. In these cases, slope values often fall between $-1$ and $-2$ or slightly lower, corresponding to smooth declines across the item spectrum. These patterns indicate that the models reflect item-level variation more clearly.

The smallest Flan-T5 model (Flan-T5 Small) displays weaker gradients in Experiment A under metalinguistic prompting, with slope values around $-0.25$ to $-0.40$. These values correspond to relatively shallow curves with limited variation across items. In Experiment B, however, the same model shows stronger gradients under some prompting conditions, suggesting that the emergence of item-level variation depends partly on the evaluation paradigm.

The Qwen models display a different pattern. In the direct probability condition, the models consistently show relatively weak gradients across both experiments. Under metalinguistic prompting conditions, however, the Qwen models behave differently depending on the experimental paradigm. In Experiment A, several conditions produce relatively strong gradients across scalar items, with slope values around $-1.5$ to $-2.2$. In Experiment B, by contrast, several Qwen conditions produce almost perfectly flat curves across the item spectrum, corresponding to slope values of zero. These results indicate that the two model families differ not only in overall accuracy but also in item-level accuracy.

\subsubsection{Experimental settings}
Finally, two experimental paradigms show the different patterns. In Experiment A, several models, particularly under metalinguistic prompting, exhibit noticeable gradients across scalar items. In Experiment B, however, some models show substantially weaker variation across items. Most notably, the Qwen models under several metalinguistic prompting conditions produce nearly flat curves, indicating that pragmatic interpretations are assigned uniformly across scalar items. This behavior results in slope values that are effectively zero. These findings suggest that the extent to which scalar diversity is reflected in model predictions depends not only on model architectures and prompting strategies but also on the evaluation paradigm used to elicit scalar interpretations.

\section{Discussion}
\subsection{Competence-Performance}
The results suggest that pragmatic inference in LLMs cannot be attributed to a single evaluation paradigm. Across models, prompting strategies, and experimental settings, neither direct probability measurement nor metalinguistic prompting consistently produced superior performance. Moreover, the graded patterns predicted under scalar diversity did not appear uniformly across conditions: while some model–condition combinations exhibited gradual gradients across scalar items, others showed relatively flat or weakly varying patterns.

From a competence–performance perspective, these findings suggest that pragmatic reasoning in LLMs emerges through an interaction between internal probabilistic representations and prompting-based responses rather than mapping cleanly onto a single evaluation method. Direct probability measurements may reflect aspects of the model’s underlying probabilistic competence, but this internal representation does not always manifest as scalar implicature behavior in isolation. Conversely, metalinguistic prompting can elicit explicit pragmatic responses, but these may reflect task-driven strategies rather than underlying knowledge. As a result, pragmatic competence cannot be reliably inferred from either evaluation paradigm alone. Instead, the observed patterns indicate that pragmatic reasoning depends jointly on various factors, such as model architectures, evaluation paradigms, and task-framing.

\subsection{Comparison with Hu \& Levy (2023)}
Compared with the findings of \citet{hu2023prompting}, several similarities and differences emerge. First, consistent with their results, direct probability measurements and metalinguistic prompting often produced divergent predictions, indicating that the two evaluation methods capture different aspects of model behavior. Second, unlike Hu and Levy’s finding that direct probability measurements generally outperform metalinguistic prompting, the present study shows no consistent performance advantage for either method. Third, consistent with prior findings, the comparison-based paradigm in Experiment B yielded higher accuracy than the single-sentence task in Experiment A. However, higher accuracy in the comparison setting does not necessarily indicate stronger pragmatic reasoning, as seen in the results of Experiment B. Finally, whereas Hu and Levy reported that prompt formats increasingly distant from the direct probability structure reduce consistency, the present results show no stable ordering among metalinguistic prompt types, suggesting that pragmatic inference is jointly shaped by model, condition, and prompt type.

\section{Conclusion}
This study examined how pragmatic reasoning in LLMs varies across evaluation methods. Using scalar implicature as a test case and scalar diversity as a graded diagnostic, this study compared direct probability estimation and metalinguistic prompting across models and conditions. The results show that pragmatic inference cannot be captured by a single evaluation paradigm: neither method consistently outperformed the other, and patterns varied across models and tasks. Moreover, graded patterns predicted by scalar diversity emerged only in certain model–condition combinations. From a competence–performance perspective, these findings suggest that pragmatic reasoning in LLMs reflects an interaction between internal probabilistic representations and prompt-based responses, underscoring the role of evaluation design.

\section*{Limitations}
This study has several limitations. First, the analysis focuses on scalar implicature as a single pragmatic phenomenon. While scalar implicature provides a well-established test case for pragmatic inference, the findings may not generalize to other forms of pragmatic reasoning. Second, the evaluation design is limited to a specific set of prompting strategies and task paradigms. Different prompt formulations or evaluation settings may lead to different patterns of model behavior. Finally, the experiments were conducted on a limited set of model families. Although both encoder–decoder and decoder-only architectures were included, the results may not fully generalize to other LLMs with different training objectives or alignment strategies.

\nocite{*}
\bibliography{latex/custom}

\newpage
\appendix
\FloatBarrier
\section{Pearson Correlations between Direct and Metalinguistic Accuracy}
\label{app:pearson}

\begin{table}[H]
\centering
\resizebox{\columnwidth}{!}{
\begin{tabular}{lccc}
\toprule
Model & MetaSimple & MetaInstruct & MetaComplex \\
\midrule
Flan-T5 Small & .20 & .20 & .24 \\
Flan-T5 Base  & .18 & .16 & .16 \\
Flan-T5 Large & $-$.10 & $-$.11 & .04 \\
Qwen 0.5B     & .20 & .07 & .08 \\
Qwen 1.5B     & .07 & .12 & .18 \\
Qwen 7B       & .24 & .22 & .27 \\
\bottomrule
\end{tabular}
}
\caption{Pearson correlations between direct and metalinguistic accuracy (Experiment A)}
\label{tab:correlation_expA}
\end{table}

\begin{table}[H]
\centering
\resizebox{\columnwidth}{!}{
\begin{tabular}{lccc}
\toprule
Model & MetaSimple & MetaInstruct & MetaComplex \\
\midrule
Flan-T5 Small & .11 & .07 & .12 \\
Flan-T5 Base  & .16 & .14 & .19 \\
Flan-T5 Large & $-$.03 & $-$.01 & $-$.04 \\
Qwen 0.5B     & $-$.15 & $-$.05 & .24 \\
Qwen 1.5B     & .19 & .11 & $-$.15 \\
Qwen 7B       & .08 & .14 & $-$.14 \\
\bottomrule
\end{tabular}
}
\caption{Pearson correlations between direct and metalinguistic accuracy (Experiment B)}
\label{tab:correlation_expB}
\end{table}

\section{Slopes of Item-Level Accuracy across Scalar Items}
\label{app:slopes}

\begin{table}[H]
\centering
\resizebox{\columnwidth}{!}{
\begin{tabular}{lcccc}
\toprule
Model & Direct & MetaSimple & MetaInstruct & MetaComplex \\
\midrule
Flan-T5 Small & $-1.52$ & $-0.278$ & $-0.248$ & $-0.392$ \\
Flan-T5 Base  & $-1.90$ & $-2.21$  & $-2.15$  & $-1.40$ \\
Flan-T5 Large & $-1.17$ & $-2.37$  & $-2.37$  & $-1.40$ \\
Qwen 0.5B     & $-0.278$ & $-2.06$ & $-1.52$  & $-0.539$ \\
Qwen 1.5B     & $-0.308$ & $-1.58$ & $-1.46$  & $-1.49$ \\
Qwen 7B       & $-0.285$ & $-2.25$ & $-2.23$  & $-1.21$ \\
\bottomrule
\end{tabular}
}
\caption{Slopes of item-level accuracy across scalar items (Experiment A)}
\label{tab:slope_expA}
\end{table}

\begin{table}[H]
\centering
\resizebox{\columnwidth}{!}{
\begin{tabular}{lcccc}
\toprule
Model & Direct & MetaSimple & MetaInstruct & MetaComplex \\
\midrule
Flan-T5 Small & $-1.52$ & $-1.15$ & $-1.34$ & $-2.05$ \\
Flan-T5 Base  & $-1.90$ & $-2.00$ & $-1.49$ & $-2.01$ \\
Flan-T5 Large & $-1.17$ & $-2.24$ & $-2.16$ & $-2.28$ \\
Qwen 0.5B     & $-0.278$ & $0$ & $0$ & $0$ \\
Qwen 1.5B     & $-0.308$ & $0$ & $0$ & $0$ \\
Qwen 7B       & $-0.285$ & $-0.003$ & $-1.91$ & $0$ \\
\bottomrule
\end{tabular}
}
\caption{Slopes of item-level accuracy across scalar items (Experiment B)}
\label{tab:slope_expB}
\end{table}

\end{document}